\documentclass{article}

\usepackage{PRIMEarxiv}

\usepackage[utf8]{inputenc} 
\usepackage[T1]{fontenc}    
\usepackage{hyperref}       
\usepackage{url}            
\usepackage{booktabs}       
\usepackage{amsfonts}       
\usepackage{dcolumn}
\usepackage{nicefrac}       
\usepackage{microtype}      
\usepackage{lipsum}
\usepackage{fancyhdr}       
\usepackage{graphicx}       
\usepackage{xcolor}
\usepackage{siunitx}
\usepackage{soul}
\graphicspath{{figures/}}

\title{FP8 Formats For Deep Learning
}

\author{
  Paulius Micikevicius, Dusan Stosic, Patrick Judd, John Kamalu, Stuart Oberman, Mohammad Shoeybi, \\ \textbf{Michael Siu, Hao Wu} \\
  NVIDIA \\
  \texttt{\{pauliusm, dstosic, pjudd, jkamalu, soberman, mshoeybi, msiu, skyw\}@nvidia.com} \\
   \And
  Neil Burgess, Sangwon Ha, Richard Grisenthwaite \\
  Arm \\
  \texttt{\{neil.burgess, sangwon.ha, richard.grisenthwaite\}@arm.com} \\
   \And
  Naveen Mellempudi, Marius Cornea, Alexander Heinecke, Pradeep Dubey \\
  Intel \\
  \texttt{\{naveen.k.mellempudi, marius.cornea, alexander.heinecke, pradeep.dubey\}@intel.com} \\
}

\begin{document}
\maketitle

\begin{abstract}

FP8 is a natural progression for accelerating deep learning training inference beyond the 16-bit formats common in modern processors. In this paper we propose an 8-bit floating point (FP8) binary interchange format consisting of two encodings - E4M3 (4-bit exponent and 3-bit mantissa) and E5M2 (5-bit exponent and 2-bit mantissa). While E5M2 follows IEEE 754 conventions for representatio of special values, E4M3's dynamic range is extended by not representing infinities and having only one mantissa bit-pattern for NaNs. We demonstrate the efficacy of the FP8 format on a variety of image and language tasks, effectively matching the result quality achieved by 16-bit training sessions. Our study covers the main modern neural network architectures - CNNs, RNNs, and Transformer-based models, leaving all the hyperparameters unchanged from the 16-bit baseline training sessions. Our training experiments include large, up to 175B parameter, language models. We also examine FP8 post-training-quantization of language models trained using 16-bit formats that resisted fixed point int8 quantization.

\end{abstract}


\section{Introduction}
Continued improvement in state of the art deep learning (DL) results has required continued increase in neural network model sizes and  compute resources needed to train them. For example, large natural language models such as GPT-3~\cite{gpt3_2020}, Turing-Megatron~\cite{megatron_2022}, PaLM~\cite{palm_2022}, and OPT~\cite{opt_2022} take weeks to train on thousands of processors. Reduced precision representation of numbers has been the corner stoen for deep learning training and inference acceleration. Common floating point types for training include IEEE single precision, TF32 mode for single precision \cite{tf32_2021}, IEEE half precision \cite{mpt2017}, and bfloat16 \cite{bfloat2020}. While some research publications have taken bit-reduction to the extreme, i.e. 1-bit binary networks~\cite{bin_conn_2015, bnn_2016, dorefa_2016, lqnet_2018, xnornet_2016}, they have not been successful in maintaining result quality needed for many practical applications.
For inference fixed-point int8 representation is a popular option. In some cases even int8 inference can encounter challenges in maintaining the accuracy required for application deployment \cite{fb_int8_accel}. Additional number representations, such as log-format~\cite{log_2016, log_2017}, posits, and log with posit exponent values \cite{log_posit_2018} have been proposed in literature but have not been adopted in practice because the demonstrated benefits have not been sufficient to justify new math pipeline hardware designs.

 FP8 is a natural progression from 16-bit floating point types, reducing the compute requirements of neural network training. Furthermore, due to its non-linear sampling of the real numbers, FP8 can also have advantages for inference when compared to int8. Wang et al. \cite{fp8_ibm_2018} proposed using 5-bit exponent format for training neural networks, confirming their methodology on the convolutional neural networks (CNNs) for image classification on CIFAR-10 and ILSVRC12 datasets. Mellempudi et al.~\cite{fp8_intel_2019} study the 5-bit exponent format for training on the larger CNNs as well as language translation networks based on recurrent and transformer blocks. Both papers investigate 16-bit weight updates as well as stochastic rounding. Use of two FP8 formats, 4- and 5-bit exponent fields, for training is introduced in~\cite{fp8_ibm_2019}, studying a wider range of CNNs as well as speech and language translation models. \cite{fp8_ibm_2019} also investigates FP8 inference of networks trained in higher precision and introduces returning of batch normalization statistics to improve result accuracy. Noune et al~\cite{fp8_gc_2022} propose a modified FP8 representation that dedicates a single encoding to special values in order to increase the represented dynamic range and present an extensive study of exponent bias effect on result quality. 8-bit inference with various formats, including FP8, with networks trained in higher precision is the focus of \cite{fp8_qcomm_2022}.
 
 In this paper we describe an 8-bit binary format for floating point representation, using two encodings for FP8. Basic principles of using FP8 for deep learning are summarized in Section~\ref{sec:dl_use}. In Section~\ref{sec:fp8_types} we describe the bit encodings and reasoning behind them. Empirical evaluation of training and inference with a variety of tasks and models is presented in Section~\ref{sec:results}. We show that FP8 training matches FP16 or bfloat16 training results for a variety of tasks and neural network model architectures and sizes, without changing any model or optimizer hyperparameters. Our study includes the training of very large language models, up to 175B parameters. It is important to consider a wide range of model sizes since it has been shown that models of different sizes may have different numerical behaviors (for example, the different behavior of Resnet-18 and Resnet-50 observed in~\cite{fp8_intel_2019}.

\section{Aspects of FP8 Usage in Deep Learning}
\label{sec:dl_use}
Some aspects of FP8 usage affect the choices for binary interchange format. For example, the dynamic ranges required by various networks dictate the need for the two formats as well as the preference for scaling factor handling in software rather than via exponent bias. Other aspects, such as type conversion specifics are orthogonal to the binary format. Both aspects are briefly reviewed in this section.

It is expected that mathematical operations on FP8 inputs will produce outputs in higher precision, optionally converting the results to FP8 prior to writing them to memory. This is common practice today for the 16-bit floating point formats (FP16 and bfloat16) found on CPUs, GPUs, and TPUs~\cite{bfloat_arm_2019,mpt2017}. For example, matrix-multiplication or dot-product instructions produce single-precision outputs but less arithmetically-intensive operations are typically performed after casting the 16-bit inputs to single precision. Thus, FP8 tensors will be generated by converting to FP8 from wider types, such as single precision floating point.

Higher precision values need to be multiplied with a scaling factor prior to their casting to FP8 in order to move them into a range that better overlaps with the representable range of a corresponding FP8 format. This is very similar to the purpose loss-scaling serves in mixed-precision training with FP16, where gradients are moved into FP16-representable range \cite{mpt2017},\cite{mpt_2018}(slides 13-16). However, some networks require per-tensor scaling factors as FP8 dynamic range is not sufficient to cover the union of all tensors' important values (see Section~\ref{sec:exponent_bias}). Details of the heuristics to select the scaling factors are beyond the scope of this paper, but the general idea is to choose a scaling factor such that the maximum magnitude in the tensor becomes close to the maximum representable magnitude in the corresponding format. Values that overflow are then saturated to the maximum representable value. Weight update skipping (and reduction of the scaling factor) on overflows, as used by FP16 automatic mixed precision training \cite{mpt_2018}, is not a good choice for FP8 as overflows are much more likely due to the narrower dynamic range, resulting in too many skipped updates. Values in higher precision get unscaled by multiplying with the inverse of the scaling factor, either after conversion from FP8 or after arithmetic instructions for a linear operation have produced a higher-precision output. In both cases only a minimal amount of additional arithmetic is required. For matrix multiplications, unscaling is applied once per dot-product, thus amortized by many multiply-accumulates with FP8 inputs. Less arithmetic intensive operations (such as nonlinearities, normalizations, or weight updates by optimizers) are typically memory-bandwidth limited and not sensitive to one additional arithmetic instruction per value.

While the mechanics of type conversion are orthogonal to the binary format, we briefly touch on some aspects for completeness of DL uses. Converting a special value from a wider precision to FP8 results in the corresponding special value in FP8. For conversions to E4M3 this means that both infinities and NaNs in the wider type (for example, single precision) turn into NaNs in FP8. This handling of special values is needed when using mixed precision training that involves both FP8 and FP16 types, since automatic mixed precision ~\cite{mpt_2018} run-time adjustment of loss-scaling relies on causing and detecting overflows. In addition, non-saturating mode of conversion can be provided for usecases that may require a strict handling of overflows. Rounding mode (round to nearest even, stochastic, etc.) choice is orthogonal to the interchange format is left up to the implementation, software and possibly hardware, for maximum flexibility.

\section{FP8 Binary Interchange Format}
\label{sec:fp8_types}
FP8 consists of two encodings - E4M3 and E5M2, where the name explicitly states the number of exponent (E) and mantissa (M) bits. We use the common term "mantissa" as a synonym for IEEE 754 standard's trailing significand field (i.e. bits not including the implied leading 1 bit for normal floating point numbers). 
The recommended use of FP8 encodings is E4M3 for weight and activation tensors, and E5M2 for gradient tensors. While some networks can train with just the E4M3 or the E5M2 type, there are networks that require both types (or must maintain many fewer tensors in FP8). This is consistent with findings in \cite{fp8_ibm_2019,fp8_gc_2022}, where inference and forward pass of training use a variant of E4M3, gradients in the backward pass of training use a variant of E5M2.

FP8 encoding details are specified in Table~\ref{tab:encodings}. We use the $S$.E.M notation to describe binary encodings in the table, where $S$ is the sign bit, E is the exponent field (either 4 or 5 bits containing biased exponent), M is either a 3- or a 2-bit mantissa. Values with a 2 in the subscript are binary, otherwise they are decimal.

\begin{table}[h]
 \caption{Details of FP8 Binary Formats}
  \centering
  \begin{tabular}{lll}
    \toprule
         & E4M3     & E5M2 \\
    \midrule
    Exponent bias & 7  & 15 \\
    Infinities     & N/A & $S.11111.00_{2}$ \\
    NaN     & $S.1111.111_{2}$  & $S.11111.\{01, 10, 11\}_{2}$ \\
    Zeros & $S.0000.000_{2}$ & $S.00000.00_{2}$ \\
    Max normal & $S.1111.110_{2} = 1.75 * 2^{8} = 448$ & $S.11110.11_{2} = 1.75 * 2^{15} = 57,344$ \\
    Min normal & $S.0001.000_{2} = 2^{-6}$ & $S.00001.00_{2} = 2^{-14}$ \\
    Max subnorm & $S.0000.111_{2} = 0.875 * 2^{-6}$ & $S.00000.11_{2} = 0.75 * 2^{-14}$ \\
    Min subnorm & $S.0000.001_{2} = 2^{-9}$ & $S.00000.01_{2} = 2^{-16}$ \\
    \bottomrule
  \end{tabular}
  \label{tab:encodings}
\end{table}

Design of these FP8 format followed the principle of staying consistent with IEEE-754 conventions, deviating only if a significant benefit is expected for DL application accuracy. Consequently, the E5M2 format follows the IEEE 754 conventions for exponent and special values and can be viewed as IEEE half precision with fewer mantissa bits (similar to how bfloat16 and TF32 can be viewed as IEEE single precision with fewer bits). This allows for straightforward conversion between E5M2 and IEEE FP16 formats. By contrast, the dynamic range of E4M3 is extended by reclaiming most of the bit patterns used for special values because in this case the greater range achieved is much more useful than supporting multiple encodings for the special values.

\subsection{Special value representations}
\label{sec:special_values}
We extend the narrow dynamic range of the E4M3 format by representing fewer special values, adopting their bit patterns for normal values. Infinities are not represented (see Section~\ref{sec:dl_use} for overflow handling details) and we retain only one mantissa bit-pattern for NaNs. This modification extents the dynamic range by one extra power of 2, from 17 to 18 binades. We gain the representation of seven more magnitudes (256, 288, 320, 352, 384, 416, 448), corresponding to the biased exponent value $1111_{2}$. The maximum representable magnitude without this modification would be 240. For consistency with IEEE 754 conventions we retain positive and negative representations for zero and NaN. While we could gain one additional representable magnitude, 480, by having just one encoding for zero and one for NaN, this would require breaking the symmetry of positive and negative representations inherent in the IEEE 754 formats, complicating or invalidating algorithm implementations that rely on this property. For example, IEEE floating point formats allow comparison and sorting of floating point values using integer operations. The benefit of increasing the maximum value from 448 to 480 for DL is not significant to warrant deviating from IEEE convention and losing software implementations that rely on it.

As mentioned earlier, E5M2 represents all the special values (infinities, NaNs, and zeros) consistently with IEEE conventions. Our extensive empirical studies (Section~\ref{sec:results}) indicate that 5 bits of exponent provide sufficient per tensor dynamic range (32 binades, including the subnormal values) for DL. Furthermore, the benefit of having fewer representations of special values would be much smaller for E5M2 than it was for E4M3 - only 3 additional magnitude values would be added due to the smaller mantissa, one additional binade is much less impactful when E5M2 already provides 32 (compared to E4M3's 17 without the adjustment). 

\subsection{Exponent bias}
\label{sec:exponent_bias}
Both E4M3 and E5M2 retain IEEE-like exponent biases: 7 and 15 for E4M3 and E5M2, respectively. Exponent bias controls the placement of representable range on the real number line. The same effect is achieved when maintaining a scale factor per tensor. Our experiments indicate that there are neural networks that cannot use the same exponent bias for all tensors of a given type, requiring a per-tensor adjustment. One such example is discussed in Section~\ref{sec:scale_granularity}. Consequently, we chose to not deviate from the IEEE convention for exponent bias. Leaving the per-tensor scaling to software implementation enables more flexibility than is possible with a programmable exponent bias approach - the scaling factor can take on any real value (typically represented in higher precision), while programmable bias is equivalent to allowing only powers of 2 as scaling factors.

\section{Empirical Results}
\label{sec:results}
Training experiments were carried out with simulated FP8 - tensor values were clipped to only those that could be represented in FP8 (including the scaling factor application and saturation). For example, prior to matrix multiplication for a fully-connected layer, both the incoming activations and the weight tensors were converted to FP8 and back to the wider representation (either FP16 or bfloat16). Arithmetic was carried out using the wider representation for two reasons: the interchange format is the focus of this paper since different processors may choose different vector- and matrix-instruction implementations, emulation of arithmetic not supported in hardware would be prohibitively slow for training of the large models. Obtaining results for large models is imperative as previous studies have identified different numerical behavior for models of different sizes (for example, R18 and R50 in~\cite{fp8_intel_2019}).

\subsection{Training}
 In the FP8 training experiments we retain the same model architectures, weight initializations, and optimizer hyper-parameters as are used for higher-precision baseline training sessions. Baselines were trained in either FP16 or bfloat16, which have been shown to match the results of single-precision training sessions \cite{mpt2017, bfloat2020}. In this study we focused on the input tensors for math-intensive operations - convolutions and matrix multiplies, to which we'll refer as GEMM-operations as they involve dot-product computations. Thus, unless otherwise specified, we clip to FP8-representable values the activation, weight, and activation gradient tensors that are inputs to GEMMs. Output tensors were left in higher precision as they typically are consumed by non-GEMM operations, such as a non-linearities or normalizations, and in a number of cases get fused with the preceding GEMM operation. Moving more tensors to FP8 is the subject of future study.
 
 \begin{table}[h]
 \caption{Image Classification Models, ILSVRC12 Validation Top-1 Accuracy}
  \centering
  \begin{tabular}{lcc}
    \toprule
    Model     & Baseline     & FP8 \\
    \midrule
    VGG-16       & 71.27  & 71.11 \\
    VGG-16 BN    & 73.95  & 73.69 \\
    Inception v3 & 77.23  & 77.06 \\
    DenseNet 121 & 75.59  & 75.33 \\
    DenseNet 169 & 76.97  & 76.83 \\
    Resnet18     & 70.58  & 70.12 \\
    Resnet34     & 73.84  & 73.72 \\
    Resnet50 v1.5 & 76.71  & 76.76 \\
    Resnet101 v1.5   & 77.51  & 77.48 \\
    ResNeXt50    & 77.68  & 77.62 \\
    Xception     & 79.46  & 79.17 \\
    MobileNet v2  & 71.65  & 71.04 \\
    DeiT small   & 80.08  & 80.02 \\
    \bottomrule
  \end{tabular}
  \label{tab:img_classification_accuracies}
\end{table}
 
 Results for image classification task are listed in Table~\ref{tab:img_classification_accuracies}. All networks were trained on ImageNet ILSVRC12 dataset, top-1 accuracy was computed on the validation dataset. All GEMM operations' inputs were clipped to FP8, including the first convolution and the last fully-connected layer, which were left in higher precision by previous studies \cite{fp8_intel_2019, fp8_ibm_2018}. DeiT~\cite{deit_2021} is a Trasnformer-based architecture, the rest of the models are CNNs. With the exception of MobileNet v2, accuracy achieved by FP8 training is within run-to-run variation of higher-precision training (run-to-run variation in achieved accuracy is observed when running training sessions initialized with different random seeds). We continue work on recovering the remaining accuracy for MobilNet v2.
 
Language translation task was tested using both Transformers and LSTM-based recurrent GNMT neural network. Even though Transformer-based translation models have superseded RNNs in practice, we include GNMT to more completely cover model architecture types, as well as a proxy for other tasks that still use recurrent networks. Models were trained on the WMT 2016 English->German dataset, evaluated using sacreBLEU on newstest2014 data (higher BLEU scores are better). Evaluation scores for FP8-trained models are within run-to-run variation variation bounds when compared to the baseline training sessions.

\begin{table}[h]
 \caption{Language Translation Models, English->German, BLEU Scores}
  \centering
   \begin{tabular}{lSS} 
    \toprule
    Model     & \multicolumn{1}{c}{Baseline}     & \multicolumn{1}{c}{FP8} \\
    \midrule
    GNMT              & 24.83 & 24.65 \\
    Transformer Base  & 26.87 & 26.83 \\
    Transformer Large & 28.43 & 28.35 \\
    \bottomrule
  \end{tabular}
  \label{tab:translation_accuracies}
\end{table}

\begin{figure}
  \centering
  \includegraphics[scale=0.7]{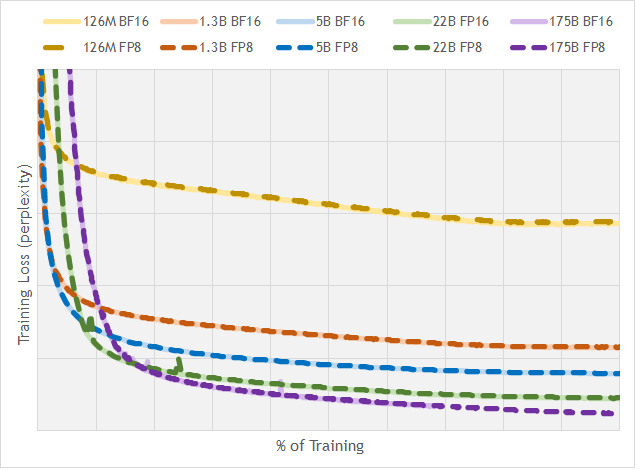}
  \caption{Training loss (perplexity) curves for various GPT-3 models. x-axis is normalized number of iterations.}
  \label{fig:gpt_curves}
\end{figure}

Training losses (perplexity, lower is better) for a variety of language models are listed in Table~\ref{tab:gpt_accuracies}. Transformer models were trained on the Wikipedia dataset. GPT models were trained on a variant of The Pile dataset~\cite{pile_2020}, augmented with Common Crawl and Common Crawl-derived datasets, as described in Section 3 of~\cite{megatron_2022}. As was seen with image networks, training results of the FP8 sessions is within run-to-run noise of 16-bit training sessions. Note that 175B parameter model perplexity is reported at 75\% training, as the bfloat16 baseline run has not yet completed. The FP8 training session has completed and its loss curve is consistent with successful training as shown in Figure~\ref{fig:gpt_curves}. As with the vision and language translation models, we conclude that FP8 training results match those of 16-bit training sessions.

\begin{table}[h]
 \caption{NLP Models, Perplexity}
  \centering
   \begin{tabular}{lSS} 
    \toprule
    Model     & \multicolumn{1}{c}{Baseline}     & \multicolumn{1}{c}{FP8} \\
    \midrule
    Transformer-XL Base  & 22.98 & 22.99 \\
    Transformer-XL Large & 17.80 & 17.75 \\
    GPT 126M     & 19.14 & 19.24 \\
    GPT 1.3B     & 10.62 & 10.66 \\
    GPT 5B       & 8.94  & 8.98  \\
    GPT 22B      & 7.21  & 7.24 \\
    GPT 175B     & 6.65  & 6.68 \\
    \bottomrule
  \end{tabular}
  \label{tab:gpt_accuracies}
\end{table}

\subsection{Inference}
8-bit inference deployment is greatly simplified by FP8 training, as inference and training use the same datatypes. This is in contrast to int8 inference with networks trained in 32- or 16-bit floating point, which require post-training quantization (PTQ) calibration and sometimes quantization-aware training (QAT) in order to maintain model accuracy. Furthermore, even with quantization aware training some int8-quantized models may not completely recover the accuracy achieved with floating point \cite{fb_int8_accel}.

We evaluate FP8 post-training quantization of models trained in 16-bit floating point. Table~\ref{tab:ptq_accuracies} lists inference accuracies for FP16-trained models quantized to either int8 or E4M3 for inference. Both quantizations use per-channel scaling factors for weights, per-tensor scaling factors for activations, as is common for int8 fixed-point. All input tensors to matrix-multiply operations (including attention batched matrix multiplies) were quantized. Max-calibration (choosing the scaling factor so that the maximum magnitude in a tensor is represented) is used for weights, activation tensors are calibrated using the best calibration chosen from max, percentile, and MSE methods. BERT language model evaluation on Stanford Question Answering Dataset shows that FP8 PTQ maintains accuracy while int8 PTQ leads to a significant loss of model accuracy. We also tried casting the tensors to FP8 without applying a scaling factor, which resulted in a significant accuracy loss, increasing the perplexity to 11.0. Evaluation of GPT models on wikitext103 dataset shows that while FP8 PTQ is much better at retaining model accuracy compared to int8.

\begin{table}[h]
 \caption{Post training quantization of models trained in 16-bit floating point. For F1 metrics higher is better, for perplexity lower is better. Best 8-bit result is bolded.}
  \centering
   \begin{tabular}{llSSr} 
    \toprule
    Model  & Dataset (metric)   & \multicolumn{1}{c}{16-bit FP} & \multicolumn{1}{c}{int8}     & \multicolumn{1}{c}{E4M3} \\
    \midrule
    BERT Base  & SQuAD v1.1 (F1) & 88.19 & 76.89 & \textbf{88.09} \\
    BERT Large & SQuAD v1.1 (F1) & 90.87 & 89.65 & \textbf{90.94} \\
    GPT3 126M  & wikitext103 (perplexity) & 19.01 & 28.37 & \textbf{19.43} \\
    GPT3 1.3B  & wikitext103 (perplexity) & 10.19 & 12.74 & \textbf{10.29} \\
    GPT3 6.7B  & wikitext103 (perplexity) & 8.51 & 10.29 & \textbf{8.41} \\
    \bottomrule
  \end{tabular}
  \label{tab:ptq_accuracies}
\end{table}

\begin{figure}
  \centering
  \includegraphics[scale=0.7]{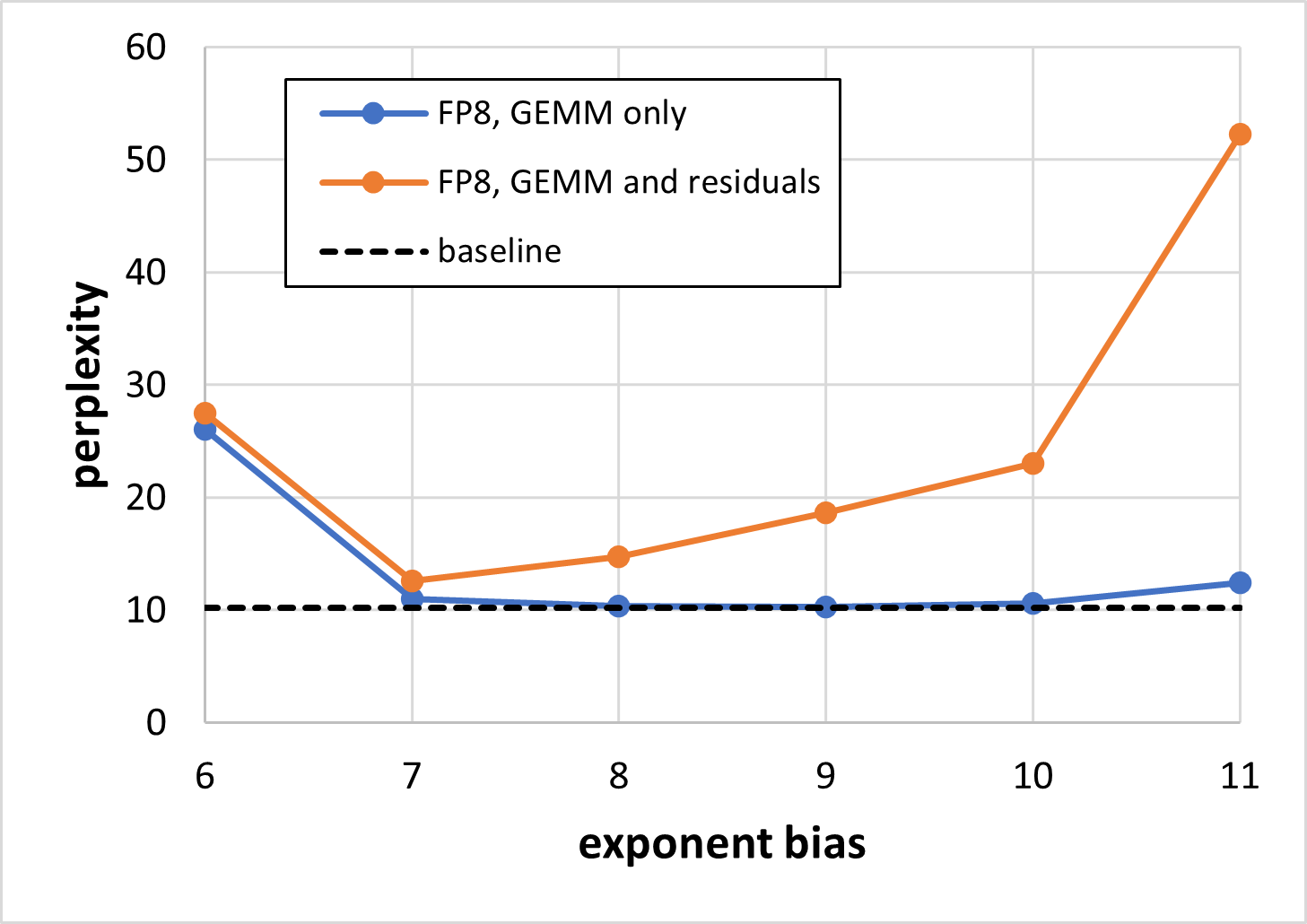}
  \caption{1.3B GPT3 perplexity when bfloat16-trained model weight and activation input tensors are cast to E4M3 format with various exponent biases, no per-tensor scaling.}
  \label{fig:gpt13_exp_biases}
\end{figure}

\subsection{Per-tensor scaling factors}
\label{sec:scale_granularity}
While training and inference for a number of networks can be successfully carried out in FP8 with the same scaling factor for all the tensors of the same type (in other words, choosing a single exponent bias could be possible), there are cases where per-tensor scaling factors are needed to maintain accuracy. This need becomes more pronounced when we store more of the tensors in FP8, not just inputs for GEMM operations. Figure~\ref{fig:gpt13_exp_biases} shows the FP8 inference perplexity (measured on wikitext103 dataset), when using post-training-quantization of a bfloat16-trained network. No calibration was done, weight and activation tensors were cast from bfloat16 to E4M3 type with the corresponding exponent bias. As we can see, when casting to FP8 only the inputs to GEMM operations (both weighted GEMMs as wells as two attention batched matrix multiplies that involve only activations), several exponent bias choices in the $[7, 10]$ range lead to results matching the bfloat16 baseline. However, if we also quantize to FP8 the residual connections (input tensors for the Add operations, which further reduces pressure on both storage and memory bandwidth) then no single exponent bias value leads to sufficient accuracy - even exponent bias of 7 results in perplexity of 12.59 which is significantly higher (worse) than 10.19 for the bfloat16 baseline. However, if instead  we calibrate the tensors to have their own scaling factors (following the convention of int8 quantization to use per-channel and per-tensor scaling factors for weights and activations, respectively \cite{int8_nv_2020}) we achieve 10.29 and 10.44 perplexities for GEMM-only and GEMM+residuals FP8 inference.

\section{Conclusions}
In this paper we propose an FP8 binary interchange format, consisting of E4M3 and E5M2 encodings. By minimally deviating from IEEE-754 conventions for binary encoding of floating point values, we ensure that that software implementations can continue to rely on such IEEE FP properties as ability to compare and sort values using integer operations. The primary motivator for the format is acceleration of Deep Learning training and inference, by enabling smaller and more power efficient math pipelines as well as reducing memory bandwidth pressure. We demonstrate that a wide variety of neural network models for image and language tasks can be trained in FP8 to match model accuracy achieved with 16-bit training sessions, using the same model, optimizer, and training hyperparameters. Using FP8 not only accelerates and reduces resources required to train, but also simplifies 8-bit inference deployment by using the same datatypes for training and inference. Prior to FP8 8-bit inference required calibrating or fine-tuning for int8 models trained in floating point, which added complexity to the deployment process and in some cases failed to maintain accuracy.

\bibliographystyle{plain}  
\bibliography{references}  

\end{document}